# Structured Proxy Features for Multimodal NSCLC Survival Prediction from Pretreatment CT

Huu Phong Nguyen[1], Delower Hossain[2], Ehsan Saghapour[3], Zhandos Sembay[3], Jake Y. Chen[1,2,3,*]

[1] Department of Biomedical Informatics and Data Science, School of Medicine, The University of Alabama at Birmingham (UAB), Birmingham, AL, United States.

[2] Department of Computer Science, The University of Alabama at Birmingham (UAB), Birmingham, AL, United States.

[3] System Pharmacology and AI Research Center (SPARC), The University of Alabama at Birmingham (UAB), Birmingham, AL, United States.

*Corresponding Author: jakechen@uab.edu

## Abstract

Lung cancer results in roughly 1.8 million fatalities annually worldwide, with non-small cell lung cancer (NSCLC) comprising the majority of cases. Despite advancements in treatment, survival stratification remains challenging due to intratumoral heterogeneity inadequately captured by conventional descriptors. Standard radiomic and deep learning techniques regard imaging features as independent quantities, overlooking structured interactions between tumor characteristics. We evaluate whether structured proxy features can enhance multimodal NSCLC survival prediction by augmenting pretreatment computed tomography (CT) representations, radiomics, and clinical variables with six simulation-derived features designed to capture interactions between heterogeneity and morphology. A radiomic-parameterized cellular automaton generates growth-rate and necrosis-ratio proxy features from baseline CT by using entropy and sphericity to compute low-dimensional proxy parameters. The imaging backbone is a Transformer-based Masked Autoencoder (TMAE), which was chosen after a systematic evaluation with alternative encoders within the same pipeline and provides attention-based visualizations that highlight tumor regions receiving higher model attention. On the public Lung1 cohort ($n = 390$), the primary four-modality fusion attained a C-index of 0.641 (iAUC 0.731, log-rank $p < 0.001$). The primary result compares favorably with prior multimodal results on Lung1 (C-index 0.631; iAUC 0.592 [15]) under a comparable evaluation protocol, while a separate exploratory coefficient-optimization analysis achieved a best observed C-index of 0.662 (iAUC 0.748). These results indicate that, in addition to conventional radiomic, deep, and clinical representations within the Lung1 benchmark, simulation-derived proxy features may provide complementary predictive information within this fixed Lung1 benchmark. By integrating structured tumor-dynamics-inspired descriptors with modern volumetric CT representations, the framework provides a practical approach for retrospective relative risk ranking from routinely acquired pretreatment imaging and establishes a foundation for future repeated-split, external-cohort, and calibration studies.

## 1. Introduction

Lung cancer is the leading cause of cancer-related mortality worldwide, with nearly 1.8 million deaths per year [1, 2]. Non-small cell lung cancer (NSCLC) constitutes approximately 85% of all lung cancer diagnoses and remains the primary subtype in clinical practice. Despite advances in treatment, the five-year survival rate is still about 28% across all stages and drops to almost 10% for patients diagnosed at stage IV [3]. Historically, prognostic models have relied on clinicopathologic factors and stage; yet, patients with analogous stage and treatment can exhibit significantly divergent outcomes, reflecting intratumoral heterogeneity that is only partially represented by standard covariates [4]. Pretreatment computed tomography (CT) is routinely obtained and contains rich phenotypic data, motivating the development of imaging-based prognostic models [5–8]. Accurate risk stratification derived from pretreatment imaging could improve prognostic ranking beyond mere staging, which has limited discriminative efficacy for individual outcomes [3, 4]. Radiomics, which is the comprehensive quantification of tumor phenotypes through high-throughput extraction of quantitative image features [5], is a noninvasive, low-cost, and repeatable approach to characterizing these phenotypic differences from routine clinical imaging.

To leverage this imaging data, radiomics [9] and deep learning [10, 14, 16-38, 62, 63] are the two essential methods to use CT to predict survival. They capture different aspects of tumor phenotype but share the same limitation: they regard quantitative imaging descriptors as independent quantities. Radiomics extracts hundreds of features that measure the intensity, shape, and texture of tumor images individually. Deep learning, on the other hand, learns volumetric representations without explicit biological structure. Yet NSCLC prognosis depends not only on characteristics of an individual tumor but also on how they interact, for instance, how intratumoral heterogeneity relates to proliferative capacity, or how boundary irregularity reflects necrosis susceptibility [41, 42, 43]. These interactions are biologically plausible and clinically relevant, but they remain unmodeled by traditional feature-extraction algorithms.

Multimodal fusion of radiomic, deep, and clinical features improves discriminative performance over unimodal approaches by combining complementary representations rather than by directly modeling these biological interactions [15]; however, performance gains due to increased architectural complexity have been minimal: on the public Lung1 cohort, diverse methods have yielded C-indices within a narrow band of 0.577–0.631 [15, 50-54]. Part of this plateau may reflect known fragilities in radiomic signatures, such as split-dependent feature selection, inter-feature redundancy, and sensitivity to extraction parameters [9]. More fundamentally, incorporating radiological imaging into multimodal fusion models remains a critical yet underdeveloped area, as radiological images offer a comprehensive three-dimensional view of tumor characteristics that other modalities cannot fully capture, but they remain difficult to integrate effectively [58]. Self-supervised learning has emerged as a viable technique for volumetric medical image representation, enabling models to learn richer CT features without requiring annotated survival labels [59]. This encourages the use of a Transformer-based Masked Autoencoder as the imaging backbone, providing a stronger representation platform than earlier autoencoder- or CNN-based imaging encoders. This stronger backbone also provides a more rigorous context for evaluating whether structured proxy features add complementary prognostic value beyond deep imaging representations alone. Here, "proxy features" refer to compact, low-dimensional prognostic descriptors generated from a radiomic-parameterized cellular automaton, including estimated growth rate and necrosis ratio, which summarize structured interactions between tumor heterogeneity and morphology from baseline CT.

In addition, cellular automata (CA) simulation frameworks [11, 12, 13, 60, 61] utilize local transition rules to model processes such as proliferation, necrosis, and competition on discrete lattices, offering a general mechanism for deriving structured features from local interactions.

In this study, we parameterize a CA using per-patient radiomic descriptors, transforming two scalar inputs (voxel intensity entropy and shape sphericity) through coupled rule-based transitions on a 3D lattice to generate six low-dimensional proxy features that provide a compact, data-driven summary of heterogeneity-morphology interactions. We expand upon the work of Ferretti and Corino [15], who integrated a 3D convolutional autoencoder with radiomics and clinical features on Lung1 (C-index 0.631). We incorporate this simulation-derived feature branch and use a Transformer-based Masked Autoencoder (TMAE) [17, 18] as the imaging backbone after controlled encoder comparison. We test the four-modality fusion of simulation-derived features, TMAE embeddings, radiomics, and clinical covariates (Figure 1) on Lung1 through ablation, encoder comparison, and sensitivity analysis. The central question is whether structured proxy features add complementary prognostic value within a multimodal NSCLC survival framework.

We have made the following contributions:

- A multimodal NSCLC survival framework that augments CT, radiomics, and clinical data with six simulation-derived proxy features designed to encode structured interactions between tumor heterogeneity and morphology.
- Controlled ablation on Lung1 demonstrating that these proxy features offer complementary prognostic value within this benchmark setting, with the primary four-modality configuration achieving a C-index of 0.641.
- An encoder comparison that selects TMAE as the imaging backbone, and sensitivity and coefficient analyses characterizing the proxy-feature branch. Source code is publicly available on GitHub.

The remainder of this paper is organized as follows. Section 2 discusses the proposed pipeline in detail, covering dataset preprocessing, the TMAE, the simulation framework, and the survival modeling strategy. Section 3 presents the experimental setup. In Section 4, we show the results, which include ablation studies, Bayesian hyperparameter search, and a comparison against state-of-the-art methods on Lung1. In Section 5 and Section 6, we discuss implications, limitations and concludes the paper.

# 2. Proposed Methods

## 2.1 Radiomic-Parameterized Simulation Framework

### *2.1.1 Simulation Dynamics*

We formulate a radiomic-guided simulation on a three-dimensional voxel lattice $\Omega \subseteq \mathbb{Z}^3$, defined by the segmented lung parenchyma. The simulator generates simulation-derived proxy features from the pretreatment CT scan that augment static radiomic descriptors (Figure 2). The simulation operates as a rule-based proxy-feature generator; the tissue-state transitions below define how entropy and sphericity propagate into the six downstream features. Each site $x \in \Omega$ carries one of five modeled tissue states, namely normal lung parenchyma L, proliferating expansion front P, modeled viable tumor state T, necrotic N, or modeled malignant subclone state M, whose evolution is governed by the following rule-based transition dynamics:

$$\partial P/\partial t = \alpha \cdot Map(x) \cdot 1_{rim}(x) - \mu P \quad (1)$$

$$\partial T/\partial t = \mu P - \gamma T - \beta \cdot T \cdot 1_{core}(x) \cdot \varphi[t \geq \tau_N] \quad (2)$$

$$\partial M/\partial t = \gamma T \quad (3)$$

$$\partial N/\partial t = \beta \cdot T \cdot 1_{core}(x) \cdot \varphi[t \geq \tau_N] \quad (4)$$

The function $1_{rim}(x)$ equals 1 for voxels in the normal lung parenchyma immediately adjacent to the tumor boundary within Ω, and 0 elsewhere; it restricts rule-based outward expansion to neighboring lung-parenchyma voxels, which upon occupation transition to state P, with each such voxel's proliferative invasion rate further scaled by the tissue-permissiveness field Map(x). Proliferating cells transition into modeled viable tumor state at rate μ and undergo malignant transition at rate γ. The function $1_{core}(x)$ equals 1 for voxels in the tumor interior and 0 elsewhere; necrosis is restricted to these interior voxels and ensues following a period of vascular insufficiency, modeled by a delay of $\tau_N$. The parameters μ, γ, and $\tau_N$ are fixed uniformly across patients; α and β are the sole per-patient quantities (Table 1).

### *2.1.2 Radiomic Parameterization*

For each of the two per-patient parameters, namely proxy proliferation coefficient α and proxy necrosis coefficient β, we select one CT-derived radiomic descriptor whose prognostic association in the NSCLC literature corresponds to the process it parameterizes. For α, we use first-order voxel intensity entropy ($E_i$), which quantifies the degree of gray-level disorder within the tumor volume. Higher entropy indicates greater spatial heterogeneity in tissue composition, a pattern associated with worse survival outcomes in NSCLC [41]; following this observation, it is used here as a parameterization anchor for proliferative behavior. For β, we use shape-based sphericity ($S_i$), which measures how closely the tumor boundary approximates a sphere. Sphericity has been confirmed as an independent predictor of recurrence and overall survival in NSCLC [42, 43]; on the premise that geometric irregularity may reflect inadequate vascular supply at the tumor core, the irregularity term $(1 - S_i)$ is used here as a parameterization anchor for necrosis susceptibility. Both descriptors are extracted from the segmented tumor volume using PyRadiomics [49] and used to define the proxy proliferation coefficient and proxy necrosis coefficient of patient *i* as the affine functions:

$$\alpha_i = a_0 + a_1 \cdot E_i, \quad \beta_i = b_0 + b_1 \cdot (1 - S_i) \quad (5)$$

In Equation (5), $E_i$ is a unit-interval normalization of the raw voxel intensity entropy $H_i$, and $(1 - S_i)$ encodes the degree of shape irregularity. The affine coefficients in Equation (5) would ideally be estimated from serial CT scans, but longitudinal imaging is rarely available in retrospective NSCLC cohorts. Patient survival is therefore used to calibrate the coefficients, under the assumption that the resulting simulation outputs may serve as additional proxy features for prognostic modeling. The four coefficients ($a_0$, $a_1$, $b_0$, $b_1$) are population-level constants shared across all patients. Two scalar features are derived from the cellular automaton simulation described in Section 2.1.1. Let $V_f = \int_\Omega [T + M](x, t_n)\, dx$ and $V_i = \int_\Omega T(x, 0)\, dx$ denote the final and initial viable tumor volumes, respectively. The growth rate and necrosis ratio are:

$$Growth\ rate = (V_f - V_i) / n \quad (6)$$

$$Necrosis\ ratio = \int\Omega N\, dx / (\int\Omega N\, dx + V_f) \quad (7)$$

These two summaries, together with $\alpha_i$, $\beta_i$, $E_i$, and $S_i$, form the six simulation-derived features $\psi_i$ used downstream. Figure 3 illustrates the simulation growth dynamics for three representative patients with distinct α and β values.

## 2.2 Transformer-based Masked Autoencoder

### 2.2.1 Architecture

TMAE is used here as the selected self-supervised imaging backbone after controlled encoder comparison (Section 4.2). TMAE is a Vision Transformer-based masked autoencoder [17, 18, 56] that learns representations by reconstructing randomly masked patches from an input volume. Its self-attention mechanism can relate spatially distant regions in a single forward pass. Here, we apply it to 3D pretreatment CT patches of lung tumors; an asymmetric encoder-decoder design ensures that only visible patches are processed by the heavier encoder, producing a 768-dimensional per-patient embedding for downstream survival modeling. A CT tumor patch of $32^3$ voxels is partitioned into N = 512 non-overlapping sub-patches of size $4^3$ voxels. Each sub-patch is flattened to a 64-dimensional vector $X_p \in \mathbb{R}^{64}$ and embedded by a learned linear map:

$$E = X_p \cdot W_{embed} + b, \quad E \in \mathbb{R}^{(512 \times 768)} \quad (8)$$

Positional encoding is used to encode the spatial position of each patch token in the sequence. The encoding is computed using sinusoidal functions factorized across the depth, height, and width axes. During TMAE training, 75% of tokens (384 of 512) are randomly masked. The encoder operates exclusively on the 128 visible tokens, reducing self-attention complexity from $O(512^2)$ to $O(128^2)$. The encoder comprises 12 transformer layers with 12 attention heads and embedding dimension d = 768 (per-head dimension $d_k = 64$). The decoder operates on 528-dimensional embeddings across 8 layers with 16 attention heads, reconstructing the complete volume from the projected encoder outputs and learnable mask tokens.

### 2.2.2 Transformer Blocks

Each transformer block applies multi-head self-attention (MHSA) followed by a position-wise feed-forward network, with layer normalization and residual connections at both sub-layers. For an input sequence X, the attention output is:

$$MHSA(X) = Concat(head_1, \ldots, head_h) \cdot W^O \quad (9)$$

where each head computes:

$$head_i = Attention(XW_i^Q, XW_i^K, XW_i^V) \quad (10)$$

The scaled dot-product attention is:

$$Attention(Q, K, V) = softmax(QK^T / \sqrt{d_k}) \cdot V \quad (11)$$

The scaling by $1/\sqrt{d_k}$ prevents inner products from growing with embedding dimension. The attention matrix A encodes pairwise token affinities, enabling the model to relate spatially distant tumor regions.

### 2.2.3 Decoder and Reconstruction Objective

For reconstruction during training, encoder representations are projected to the decoder embedding space by a learned linear map $W_{proj} \in \mathbb{R}^{(768 \times 528)}$:

$$x_{dec} = x_{enc} \cdot W_{proj} + b \quad (12)$$

The decoder operates on the projected visible tokens together with learnable mask tokens $m \in \mathbb{R}^{(384 \times 528)}$ standing in for the withheld positions. A linear prediction head maps each decoder token back to the sub-patch dimension $p^3 = 64$ for voxel-intensity reconstruction. The reconstruction objective $\mathcal{L}_{TMAE}$ is the mean squared error over masked positions:

$$\mathcal{L}_{TMAE} = (1/|M|)\, \boldsymbol{\Sigma}_{i \in M} \|y_i - \hat{y}_i\|^2 \quad (13)$$

Restricting the loss to masked positions encourages the encoder to learn representations that generalize across the tumor volume rather than memorizing local voxel patterns. Following training, a learnable CLS token prepended to the sequence serves as the global tumor descriptor; the final-layer CLS embedding $z_i \in \mathbb{R}^{768}$ is extracted at inference time with all 512 spatial tokens processed by the encoder. This embedding provides an off-the-shelf volumetric representation of the tumor that can be used directly for downstream survival prediction without end-to-end fine-tuning of the encoder.

## 2.3 Feature Integration and Survival Prediction

Survival modeling is a key component of the proposed framework, serving both to calibrate the simulation parameters α and β against patient outcomes and to receive, as input, the multimodal fusion capturing complementary facets of NSCLC. The four feature groups (simulation-derived features $\psi_i \in \mathbb{R}^6$, TMAE embeddings $z_i \in \mathbb{R}^{d_z}$, conventional radiomics $\rho_i$, and clinical covariates $c_i$) are concatenated as $f_i = [\psi_i \parallel z_i \parallel \rho_i \parallel c_i]$ and passed to two complementary survival models, CoxPH and Neural Cox, whose outputs are combined into an ensemble risk score.

### *2.3.1 Survival Models*

Two survival models are trained on the fused representation $f_i$, each expressing the individual hazard as a function of $f_i$ relative to a common baseline hazard $h_0(t)$.

**Cox Proportional-Hazards (CoxPH) [44].** The log-hazard ratio is a linear function of $f_i$:

$$h_i^c(t) = h_0(t) \cdot exp(\beta_c^T f_i) \quad (14)$$

where $\beta_c$ is estimated by maximizing the Cox partial log-likelihood. The risk score is $r_i^c = \beta_c^T f_i$.

**Neural Cox [45].** To capture non-linear interactions among the fused features, the log-hazard ratio is parameterized by a multilayer perceptron $g_\theta: \mathbb{R}^{d_f} \to \mathbb{R}$:

$$h_i^n(t) = h_0(t) \cdot exp(g_\theta(f_i)) \quad (15)$$

trained by minimizing the negative Cox partial log-likelihood:

$$\mathcal{L}(\theta) = - \Sigma_{i:\delta_i=1} (g_\theta(f_i) - log\ \Sigma_{j \in \mathcal{R}(t_i)} exp(g_\theta(f_j))) \quad (16)$$

### *2.3.2 Ensemble Risk Score*

The final patient risk score is obtained by averaging the two min-max normalized scores:

$$r_i = ½ \cdot \sigma(r_i^c) + ½ \cdot \sigma(r_i^n) \quad (17)$$

where σ(·) denotes min-max normalization using the minimum and maximum of the training fold, applied unchanged to the held-out fold. Combining linear and non-linear scores reduces sensitivity to the functional form of the risk–feature relationship.

# 3. Experiments

## 3.1 Benchmark Dataset

We used the NSCLC-Radiomics ("Lung1") cohort available through the Cancer Imaging Archive (TCIA; DOI: 10.7937/K9/TCIA.2015.PF0M9REI) [39], comprising an original cohort of 422 non-small cell lung cancer patients (stage IA–IIIB) treated at the MAASTRO Clinic, Netherlands. The dataset was originally de-identified under the Cancer Genome Atlas (TCGA) protocol and made publicly available without restriction; accordingly, this study was exempt from institutional review board approval. Each subject underwent pretreatment CT imaging acquired at a resolution of 512×512 pixels across a variable number of slices with a voxel size of approximately 1×1×1 $mm^3$, and the gross tumor volume (GTV) was manually delineated by a radiation oncologist and provided as a binary segmentation mask. After excluding 32 cases with missing data, the analysis cohort comprised 390 patients, of whom 345 experienced the event (88.5% event rate), with a median survival of 18.1 months; staging distribution was 21.3% stage I, 9.2% stage II, and 69.2% stage III, and 0.3% stage unknown. Overall survival time and censoring status were used as the primary endpoint throughout all experiments. Cohort characteristics are summarized in Table 2. The available clinical covariates used as model inputs included age, sex, and clinical stage variables. Treatment-related variables, such as chemotherapy regimen, radiation dose, or surgical approach, and molecular markers, such as EGFR mutation status, remain excluded from the public Lung1 dataset and were therefore not incorporated into the model.

## 3.2 Data Split

For consistency with the closest preceding multimodal Lung1 benchmark, Ferretti and Corino [15], we adopted a single fixed-split evaluation protocol, which provides a matched basis for the relative comparison provided in Section 4.1. Patients were partitioned into training and test sets using a stratified 2/3 split to preserve the event-rate distribution. The 390-patient analysis cohort was divided into 260 patients for training and 130 patients for testing. All model selection, feature selection, and hyperparameter tuning were performed within the training set using cross-validation, including 10-fold cross-validation for radiomic feature stability and 10-fold cross-validation for survival-model tuning. All experiments utilized the same fixed split and were conducted independently; the feature-selection search in Section 4.4.1 and the coefficient optimization in Section 4.4.3 were distinct studies with separate trial budgets.

## 3.3 Preprocessing

All CT volumes were resampled to 1-mm³ isotropic resolution (trilinear interpolation), windowed to a lung-appropriate intensity range (e.g., −1000 to 400 HU), and linearly scaled to [0,1]. Tumor regions of interest (ROIs) were cropped around the provided segmentation masks for radiomic extraction and self-supervised pretraining. For TMAE training and inference, we extracted tumor-centered cubic patches (32×32×32 mm³); tumors exceeding patch bounds were center-cropped, and smaller tumors were zero-padded with padding masks excluded from the reconstruction loss.

## 3.4 Radiomics Feature Extraction

We extracted radiomic features from the ROI using PyRadiomics [49]. Specifically, we computed 1,688 features with PyRadiomics (v3.0.1): shape (n=14), first-order intensity (n=18), and texture descriptors (GLCM, GLRLM, GLSZM, GLDM, NGTDM). Image pre-processing included Laplacian-of-Gaussian filtering ($\sigma \in \{1.0, 2.0, 3.0, 4.0, 5.0\}$ mm) and wavelet decomposition (8 sub-bands). To reduce redundancy and improve robustness, we performed feature selection via 10-fold cross-validation on the training set: in each fold we removed constant features and correlated features, and retained features selected in a majority of folds, exceeding a consensus threshold.

## 3.5 Hyperparameter Setup Details

The following summarizes the general configuration shared across all experiments; experiment-specific settings are described in the corresponding subsections of Section 4 and the Supplementary Material. Full hyperparameter specifications for all components are provided in Table 3.

# 4. Results

The results are organized across four subsections. Section 4.1 compares against published methods on Lung1. Section 4.2 compares three imaging encoders across progressive simulation feature configurations. Section 4.3 ablates modality arms to isolate each component's contribution; the four-modality fusion result from this experiment serves as the primary benchmark result. Section 4.4 consolidates three additional experiments: Bayesian feature-selection hyperparameter optimization, sensitivity analysis of simulation parameters, and exploratory coefficient optimization reaching a best C-index of 0.662.

## 4.1 Comparison with State-of-the-Art

Since the TCIA Lung1 dataset became publicly available in 2014 [39], survival prediction in NSCLC has drawn significant attention from the research community. The cohort offers expert GTV delineations, (Digital Imaging and Communications in Medicine) DICOM CT volumes, and right-censored overall survival data in a single standardized repository, making it the default open benchmark for radiomic survival models. A comparison with state-of-the-art methods is summarized in Table 4.

Welch et al. [50] approached the problem from a validation standpoint, applying Cox regression with a compact radiomic signature to demonstrate how feature selection leakage, signature instability, and overfitting to small splits each independently inflate apparent performance. The methodological contribution is substantial: by isolating these failure modes, the study explains why radiomic survival models so often fail to replicate across cohorts. The C-index of 0.600 reflects principled restraint rather than a serious attempt at discriminative performance and is best understood as a lower bound on what a properly validated radiomic model can achieve on Lung1.

Haarburger et al. [51] asked whether learned image representations could substitute manually engineered radiomic descriptors, combining a 2D patch-level (Convolutional Neural Network) CNN with handcrafted features as a binary survival classifier. The conceptual strength is that the network learns task-relevant spatial patterns without manual feature specification, and the combination outperformed either modality alone (C-index 0.623). The structural limitation is that 2D patches discard volumetric continuity, such that prognostically relevant inter-slice relationships are not recoverable from independent 2D views, leaving room that 3D representations subsequently exploited.

Haarburger et al. [52] shifted focus from prediction accuracy to feature reliability, using a probabilistic U-Net to retain only stable radiomic features. The robustness-first philosophy addresses a genuine clinical concern, given that radiomic features vary substantially with small changes in tumor delineation. The practical cost was a C-index of 0.577: filtering for stability removed features that, while segmentation-sensitive, still carried real prognostic information, illustrating the tension between reproducibility and discriminative performance.

Scalco et al. [53] applied a genetic programming (GP) framework that constructs composite features by combining morphological and texture descriptors through arithmetic operations, with the aim of discovering non-linear interactions that manual engineering misses. The tree-structured output is in principle interpretable, and the approach is notable for generating features rather than selecting them. The gain was nonetheless modest (C-index 0.608), suggesting that arithmetic composition of existing radiomic descriptors does not fully recover the prognostic structure latent in the imaging data.

Ferretti and Corino [15] represent the closest methodological precedent to the present work, combining PyRadiomics features with 3D convolutional autoencoder latent embeddings and clinical variables under Cox regression with LASSO penalization. The multi-domain fusion is well-motivated: handcrafted features encode known biological correlates, the autoencoder captures spatial patterns beyond manual description, and clinical variables contextualize the imaging. The result (C-index 0.631, $p$ = 0.031) is the strongest prior multimodal performance on Lung1 under a comparable evaluation protocol. The limitation is that the autoencoder encodes morphology and texture from a single pretreatment scan but captures nothing about growth dynamics or proliferative behavior, as the latent space is static by construction.

Paolo et al. [54] trained an EfficientNetB0 backbone with a soft-attention mechanism to weight CT slices by prognostic relevance, providing clinically legible spatial explanations without manual feature extraction. The attention mechanism is an interpretability strength absent from most prior approaches. The C-index of 0.584, however, suggests that attending to informative 2D slices does not substitute for modeling the tumor volumetrically, given that prognostic information is distributed across the volume and inter-slice relationships are not recoverable from a weighted average of independent 2D views.

These approaches share a notable pattern: despite advancements from hand-crafted Cox models through 2D CNNs, stability-filtered radiomics, multimodal (Autoencoder) AE, and attention networks, the C-index on Lung1 has moved within a narrow band of 0.577–0.631. Our proposed method introduces a radiomic-parameterized cellular automaton simulation, where entropy and sphericity drive the growth and necrosis parameters, to generate simulation-derived features that complement static radiomic descriptors. When fused with TMAE embeddings, PyRadiomics descriptors, and clinical variables, the primary four-modality benchmark yields a C-index of 0.641. In a separate coefficient-optimization study (Section 4.4.3), the best exploratory configuration reaches 0.662 (95% CI: [0.614, 0.719]).

## 4.2 Comparative Evaluation of Imaging Encoders: TMAE, CAE, and MedicalNet-RN18

In this section we compare our proposed pipeline using TMAE against two other deep learning models: AE and CNN, with the goal of selecting the most suitable imaging backbone for the downstream multimodal survival pipeline. To isolate the effect of encoder choice, all three encoders are evaluated within the same multimodal fusion framework, with each encoder projected to a common 768-dimensional latent space. For the Convolutional Autoencoder (CAE), we use the same configuration as TMAE, as the two share similar design principles: 3D tumor volumes are used to train the model through self-supervised masked reconstruction, where the model learns to reconstruct withheld patches from visible context, and the resulting latent representations are extracted from the bottleneck and fused with the remaining pipeline components. For the CNN model, since it operates in a supervised learning manner [57], we use pretrained weights for medical images and extract features from its last layer, using ResNet18 initialized with MedicalNet-RN18 weights, hereafter referred to as RN18. In addition, we also vary the number of simulation features incorporated into the pipeline across five progressive configurations (A1–A5): the encoder-alone model (A1), the two radiomic scalars (E and S) added (A2), the proxy coefficients (α and β) added (A3), the two simulation outputs (growth rate and necrosis ratio) added (A4), and the complete six-feature (Simulation) SM set (A5). Full implementation details are provided in Supplementary Material Section S2.

As we can glean from Figure 4, which presents the forest plots for TMAE, CAE, and RN18, several consistent patterns emerge across all three encoders.

For TMAE (Figure 4A, top panel), the encoder-alone model (A1) yields C-indices of 0.608, 0.603, and 0.597 under CoxPH, Neural Cox, and Ensemble respectively. Adding the radiomic scalars entropy and sphericity (A2) produces gains under CoxPH (0.615) and Neural Cox (0.620) but a decrease under Ensemble (0.587). Incorporating the proxy coefficients α and β (A3) yields similar improvements (0.616, 0.623, 0.649), with the Ensemble model showing a notably stronger response. Adding the simulation outputs growth rate and necrosis ratio (A4) maintains comparable performance under CoxPH (0.605) while continuing to improve Neural Cox (0.631) and Ensemble (0.653). The full six-feature SM set (A5) achieves the highest Neural Cox C-index of 0.636, though the Ensemble result (0.608) is slightly below A4. Overall, adding simulation-derived features shows more pronounced improvement under Neural Cox and Ensemble than under CoxPH. For TMAE, the pair of simulation outputs, specifically growth rate and necrosis ratio (A4), performs slightly better than the proxy coefficients α and β (A3), as evidenced for example by the Ensemble C-index of 0.653 versus 0.649.

Similar patterns are observed for CAE and RN18, where Neural Cox and Ensemble generally outperform CoxPH across all feature configurations. In addition, the simulation output pair of growth rate and necrosis ratio (A4) performs better than the other feature pairs in most cases, consistent with the TMAE findings. Among the three imaging backbones, TMAE yielded the highest downstream C-index of 0.653, compared with 0.642 for CAE and 0.627 for RN18. The gain over CAE is modest but consistent across configurations, while the gap over RN18 is larger. All three were evaluated within the same fusion framework and output space, directly supporting backbone selection for the present pipeline. TMAE was therefore selected as the imaging backbone for the subsequent multimodal analyses, while RN18 was retained only as the internal development baseline for tracking the effect of later multimodal extensions. One possible explanation for TMAE's stronger downstream performance is that its self-attention mechanism (Section 2.2.1) may better preserve distributed heterogeneity patterns across the tumor volume in this pipeline.

The relative ordering of Neural Cox and CoxPH across configurations reveals a further consistent pattern. At A1, Neural Cox trails CoxPH for all three backbones (TMAE: 0.603 vs 0.608). This ordering reverses progressively as simulation-derived features are added, with the most pronounced shift occurring between A3 and A4, when growth rate and necrosis ratio enter the pipeline, rather than between A1 and A2, when the raw radiomic scalars entropy and sphericity are first introduced. This pattern is consistent across TMAE, CAE, and RN18.

To further characterize the selected TMAE encoder, CLS-to-patch attention weights were extracted from the final encoder block, averaged across 12 heads, and projected onto the CT isosurface for representative test patients (Figure 4B). For LUNG1-014, a high-risk patient with a branching tumor morphology, attention concentrates at the morphological junctions of the tumor structure, indicating that the encoder identified spatially specific regions as most informative. For LUNG1-046, a low-risk patient with a large homogeneous tumor filling the full patch, attention distributes uniformly across all 512 tokens without any spatially identifiable structure. These two cases suggest that the TMAE encoder captures spatially distributed heterogeneity when tumor morphology is complex and falls back to global volumetric encoding when it is not, which is consistent with the self-attention mechanism described in Section 2.2.1 and may partly account for its stronger downstream performance relative to CAE and RN18.

### 4.3 Multimodal Feature Fusion Ablation Study

To isolate the contribution of each modality arm, we conducted a controlled ablation study using TMAE as the anchor component and progressively incorporating SM, Radiomics (R), and Clinical (C), with results summarized in Table 6. Starting from TMAE (C-index = 0.608 under CoxPH), adding SM (TMAE+SM) yields a notable gain under Neural Cox (0.636, $p < 0.001$) and produces the highest HR of 2.31, indicating a strong separation between risk groups. The combination of all four arms (TMAE+SM+R+C) achieves the highest overall C-index of 0.641 (Ensemble, $p < 0.001$, iAUC = 0.731). Notably, TMAE+SM (six features) outperforms TMAE+R (hundreds of selected radiomic features), suggesting that low-dimensional structured proxy features may capture prognostically useful interactions not fully recovered when radiomic descriptors are treated independently.

The proposed pipeline comprises 113.07M parameters in total, which is typical among deep learning models of similar scope. All experiments were conducted on a high-performance computing cluster equipped with one NVIDIA A100 GPU, an Intel Xeon Gold 6126 CPU (2.60 GHz), 14 allocated CPU cores, and 224 GB RAM.

## 4.4 Additional Experiments

This section consolidates three additional experiments that characterize specific components of the proposed framework: Bayesian feature-selection hyperparameter optimization (Section 4.4.1), sensitivity analysis of simulation parameters (Section 4.4.2), and exploratory coefficient optimization (Section 4.4.3).

### 4.4.1 Bayesian Feature Selection and Model Optimization

In this experiment, we analyze the effect of feature size and selection strategy on survival prediction performance; the primary multimodal evidence is presented separately in Section 4.3. We conduct a Bayesian hyperparameter search using the Tree-structured Parzen Estimator (TPE) [48] sampler across four modality arms: TMAE, Radiomics, SM, and Clinical. Specifically, five hyperparameters were jointly optimized: the TMAE correlation filtering threshold (TMAE Corr %), the TMAE selection frequency threshold (TMAE Sel %), the number of top Radiomics features (R Feat), the SM feature retention percentage (SM %), and the Clinical feature retention percentage (Clin %). The resulting total number of features (Tot Feat) is a derived diagnostic quantity reported alongside these. Motivated by the findings of Ferretti and Corino [15], we restrict the number of Radiomics features to fewer than 100. A total of 700 Bayesian trials were conducted. Figure 5 presents a parallel coordinates plot illustrating how each hyperparameter configuration maps to the final C-index across all trials for the Neural Cox model, with green indicating higher-performing configurations. Several consistent patterns emerge from the high C-index trials:

- **TMAE Corr %:** green lines cluster tightly around 60–65%, suggesting a low correlation threshold is optimal for TMAE feature filtering, while higher thresholds are associated with lower C-index.
- **TMAE Sel %:** green lines spread across 50–70% with a slight concentration around 60%, indicating moderate selection frequency is preferred.
- **R Feat:** green lines converge at around 75, suggesting that a smaller, compact Radiomics feature set performs better.
- **SM %:** green lines concentrate from 0–70%, suggesting moderate SM retention is sufficient and very high retention does not further improve performance.
- **Clin %:** green lines cluster strongly at 80–100%, indicating that retaining most or all Clinical features is critical for high performance.
- **Tot Feat:** green lines concentrate around 250–350 total features, with very low or very high feature counts consistently associated with lower C-index.

### 4.4.2 Sensitivity Analysis of Simulation Framework Parameters

To assess the robustness of the simulation module to uncertainty in its calibrated simulation parameters, we conducted a sensitivity analysis by systematically perturbing the proxy proliferation coefficient $\alpha$ and the proxy necrosis coefficient $\beta$ independently and jointly at ±10% and ±20% of their calibrated values. For each of the twelve perturbation scenarios, the full cellular automaton was re-run from scratch for the test set using the perturbed parameter values, producing updated simulated growth rate and necrosis ratio that were then propagated through the complete feature extraction and risk prediction pipeline. Four metrics were computed for each scenario relative to the unperturbed baseline: mean absolute risk score deviation ($\Delta$Risk), Spearman rank correlation ($\rho$) between perturbed and baseline risk rankings, percentage of patients whose high/low risk classification changed (Category Chg), and change in concordance index (ΔC-index).

The results demonstrate that the pipeline output is numerically stable with respect to perturbations in the calibrated values of α and β (Table 5). It should be noted that this analysis establishes numerical stability of the surrogate features, not biological validity of the simulation parameters or their correspondence to actual tumor growth. Perturbations to the proxy proliferation coefficient α produced the larger effects, with ΔRisk increasing monotonically from 0.073–0.091 at ±10% to 0.135–0.184 at ±20%, and patient reclassification reaching a maximum of 7.7% at α −20%. Nevertheless, Spearman rank correlations remained at or above 0.989 across all α scenarios, and ΔC-index was confined within |ΔC-index| ≤ 0.007. Perturbations to the proxy necrosis coefficient β produced consistently smaller effects than α across all perturbation levels, with ΔRisk ranging from 0.023 to 0.030 and a maximum reclassification of 3.1%, reflecting the comparatively limited contribution of β to the overall risk score relative to the proliferation-driven growth trajectory. The combined α+β perturbations yielded results close to α alone in all cases.

A directional asymmetry is evident within the α perturbation scenarios. Positive perturbations consistently produced larger ΔRisk than negative ones at the same level (+10%: 0.091 vs −10%: 0.073; +20%: 0.184 vs −20%: 0.135). Patient reclassification followed the opposite asymmetry, peaking at α −20% (7.7%) rather than α +20% (6.2%), suggesting that downward shifts in α displace a marginally larger proportion of patients across the risk boundary despite producing smaller absolute score deviations.

Full implementation details are provided in Supplementary Material Section S3.

### 4.4.3 Exploratory Coefficient Optimization

In this exploratory experiment, we optimize the affine scaling coefficients $a$ and $b$ in the formulas for α and β according to survival data by varying $a \in \{0.02, 0.10, 0.20\}$ and $b \in \{0.05, 0.15, 0.40\}$, using 100 Bayesian TPE trials per combination across all pipelines of the proposed framework.

Figure 6 presents a heatmap of the best C-index per α/β formula combination, with each cell reporting the C-index, best model type, and selected feature counts. Neural Cox was the top-performing model across all nine combinations. The $b = 0.05$ row consistently yielded the strongest performance across all $a$ values (0.658, 0.641, 0.662). The global best combination (dashed border) was identified at $a = 0.20$, $b = 0.05$, yielding a C-index of 0.662 under Neural Cox with SM = 2, TMAE = 63, R = 75, C = 6, and a total of 146 features. Under this configuration, the Neural Cox model yielded a C-index of 0.662 (95% CI: [0.614, 0.719]; iAUC = 0.748, log-rank $p < 0.001$, HR = 2.21).

## 5. Discussion

This study evaluates whether structured proxy features derived from baseline CT descriptors improve multimodal NSCLC survival prediction within a fixed benchmark framework. The main finding is that a compact rule-based proxy-feature branch adds prognostic value beyond standard deep, radiomic, and clinical inputs on Lung1. The primary four-modality fusion achieved a C-index of 0.641, indicating improved performance relative to the closest prior multimodal benchmark on Lung1 [15], while remaining within a single-benchmark evaluation setting. A separate exploratory coefficient-optimization analysis achieved a best observed C-index of 0.662, suggesting additional potential under tuned simulation-parameter configurations but not serving as the primary benchmark result. Simulation-derived features contribute consistent prognostic value across all three tested imaging encoders, suggesting that they may capture prognostically useful structure not well represented when radiomic and deep features are treated independently. The evidence presented here supports the narrower finding that structured proxy features add prognostic value within this specific multimodal pipeline on Lung1, where the model demonstrates discriminatory ranking rather than calibrated probability estimation or threshold-validated risk stratification, though this remains a single-benchmark observation pending stability testing across alternative splits and external cohorts.

Despite the modest cohort size, each case contributes a full volumetric CT scan, making compact self-supervised 3D representations a reasonable design choice for the imaging backbone; consistently, the superiority of Neural Cox over CoxPH across experiments suggests that the relationship between the multimodal feature set and survival is non-linear.

The configuration-dependent response of Neural Cox in Figure 4 extends the observation that Neural Cox consistently outperforms CoxPH across experiments. The advantage is not uniform: Neural Cox begins below CoxPH at A1 and gains most substantially between A3 and A4, coinciding with the addition of growth rate and necrosis ratio, which are outputs derived through coupled cellular automaton transitions rather than extracted directly from imaging. This is consistent with the interpretation that Neural Cox may better capture non-linear interactions among simulation-derived outputs, and may partly explain why the relationship between the multimodal feature set and survival appears non-linear within this pipeline.

A notable finding from the ablation study is that TMAE+SM consistently outperforms TMAE+R, despite SM comprising only six features compared to several hundred selected radiomic features. Two possible explanations are that the rule-based coupling of entropy and sphericity produces a compact prognostic summary not recoverable from independent radiomic descriptors, and that high-dimensional radiomic feature sets may suffer from redundancy or instability that dilutes their marginal contribution when combined with deep representations. The apparent advantage of the six-feature SM arm may reflect both structured summarization and residual fragility in high-dimensional radiomic integration. Resolving which explanation predominates, whether genuinely useful structured coupling or radiomic integration instability, is a key open question for future work on this pipeline. Joint-Embedding Predictive Architectures (JEPA) [55], which predict target representations in embedding space rather than reconstructing raw voxel values, offer a complementary self-supervised paradigm that may extend the masked reconstruction strategy used by the TMAE.

Entropy and sphericity are used here as interpretable parameterization anchors with plausible biological correspondence, not as direct validated measurements of proliferation or necrosis. Entropy has been independently associated with intratumoral heterogeneity and worse prognosis in NSCLC [41, 43], consistent with the observation by Aerts et al. [5] that features quantifying intratumoral heterogeneity have strong prognostic power across cancer types. Sphericity reflects tumor boundary regularity, which is linked to invasive morphology. These associations provide face validity for the mapping but do not constitute biological calibration. The sensitivity analysis confirms that the pipeline is numerically stable around the chosen parameterization; however, the perturbation analysis supports numerical robustness of the proxy features under local parameter changes, not evidence of biological correctness.

The asymmetric response to positive and negative perturbations of $\alpha$ warrants brief consideration. The larger $\Delta$Risk under positive perturbations is consistent with the nonlinear dynamics of the cellular automaton, where higher proliferation rates accelerate expansion into neighboring voxels and compound across iterations, amplifying small upward changes in $\alpha$ more than equivalent downward changes. The opposing asymmetry in reclassification, which peaks under negative perturbation, suggests that risk scores near the median split boundary are more sensitive to reductions in proliferative drive than to equivalent increases, though the overall proportion of reclassified patients remains modest across all scenarios. Taken together, these findings support the numerical stability of the simulation branch while highlighting that $\alpha$, as the primary driver of growth trajectory, contributes more to risk score variability than $\beta$ within this pipeline.

The following table (Table 7) summarizes the biological assumptions, rationale, and scope of each parameterization choice.

*Biological scope.* The simulation branch should be interpreted as a data-driven, phenomenological transformation of baseline imaging descriptors into prognostic proxy features, not as a validated reconstruction of patient-specific tumor biology. The present study does not validate the simulator against serial tumor measurements, pathology-derived necrosis or proliferation indices, or treatment-response trajectories. True biological validation would require testing the simulation outputs against serial imaging, histopathological proliferation and necrosis measures, or longitudinal treatment-response data, which are not available in the present cohort.

Compared to the closest prior work [15], which combined a 3D convolutional autoencoder with radiomics and clinical features under CoxPH with LASSO penalization, the proposed framework introduces two key differences: the replacement of the CAE with the TMAE, and the addition of a radiomic-parameterized simulation. In addition to discriminative performance, the self-attention mechanism of TMAE yields CLS-to-patch attention weights that can be projected directly onto the tumor volume, providing a degree of spatial explainability not available from the convolutional bottleneck of CAE or the global average pooling of RN18 without additional post-hoc analysis.

The proposed framework requires only pretreatment CT and standard clinical variables, both of which are routinely collected in oncology practice, and does not depend on additional invasive assays, genomic profiling, or high-cost imaging modalities. The modest improvement in discriminative performance is achieved using routinely acquired imaging, while the simulation-derived proxy features remain auditable, as each feature traces directly to two named radiomic descriptors and a defined set of CA transition rules. The present results demonstrate retrospective risk-stratification feasibility on a single public benchmark. Although the feature-selection study (700 trials) and coefficient-optimization study (900 trials) are conducted independently, the total optimization volume on a single cohort warrants caution, reinforcing the need for external validation. The most direct next step is evaluation on an independent public NSCLC cohort with calibration and threshold assessment.

*Potential clinical use case.* A plausible long-term application is baseline prognostic stratification from routinely acquired pretreatment CT. The model produces a continuous relative risk score that could be evaluated in future studies for its ability to rank patients by outcome risk. Because the present study evaluates discrimination on a single benchmark cohort without calibration, threshold validation, or cross-site testing, any workflow use remains speculative and would require substantial further evaluation. The integrated time-dependent AUC is evaluated over a 3- to 24-month window; the Cox-derived risk score reflects relative patient ordering under proportional hazards and does not produce calibrated time-specific survival probabilities. A summary of input, output, intended use, and constraints is provided in Table 8.

*Requirements before deployment.* These requirements represent substantial unresolved translational steps. A realistic deployment pathway would require: (i) standardized or automated tumor segmentation to replace manual GTV delineation; (ii) locked feature extraction and model weights to prevent post-deployment drift; (iii) calibration on local imaging protocols, as radiomic features are sensitive to scanner and reconstruction parameters; (iv) prospective site-specific validation; and (v) validation of clinically meaningful decision thresholds and risk-group stability across sites. Until these steps are completed, the framework is intended for retrospective decision support and hypothesis generation only.

*Limitations.* There are several limitations to acknowledge. Most importantly, all analyses are reported on a single fixed train/test split of one public cohort, so the stability of the observed gains across alternative partitions and independent datasets remains to be established. Repeated-split and external-cohort validation will be crucial to assess the robustness and generalizability of the observed improvement, including relative to prior benchmarks such as Ferretti and Corino [15]. Initially, the simulation develops from a single pretreatment scan and generates feature proxies rather than predictions of real tumor growth. The affine coefficients used in the simulation were calibrated using patient survival rather than observed tumor-growth trajectories since longitudinal CT imaging was not available in the Lung1 cohort. This is an important limitation because the simulation-derived features should be interpreted as prognostic proxy descriptors rather than biologically validated estimates of patient-specific tumor progression. Future studies using longitudinal imaging data will be needed to determine whether these features correspond to actual tumor-growth dynamics [40]. A practical strength of the approach is that it requires only a single baseline CT scan, which makes the framework applicable to retrospective cohorts where serial imaging is unavailable. Second, the study is based on a single publicly available cohort (Lung1, n = 390) with a high event rate (88.5%), and a more balanced or larger cohort would enhance generalization and reliability of the reported results. Third, the methodology has not been tested on an external independent validation set; the results should therefore be regarded as preliminary and subject to further validation. Additionally, all radiomic features, TMAE embeddings, and simulation-derived features in this study were extracted from manual GTV delineations provided in the public Lung1 dataset. Manual segmentation introduces inter-observer variability that may affect both radiomic and deep learning outputs, and the stability of the pipeline under alternative delineations or automated segmentation has not been tested. Fourth, treatment-related variables and molecular markers are not available in the public Lung1 release and were thus not incorporated into the model. These variables may represent important prognostic confounders, as survival can be influenced by differences in treatment exposure and tumor biology beyond imaging phenotype alone. Future studies with access to treatment records, molecular profiling, or genomic data could evaluate the incremental prognostic value of these variables alongside imaging-derived features. Fifth, imaging protocol and scanner variability were not evaluated. The Lung1 dataset was collected retrospectively at a single site and released publicly without individual-level acquisition metadata; given the known sensitivity of radiomic features to scanner and reconstruction parameters, multi-site or multi-protocol validation would be needed to assess cross-scanner reproducibility.

## 6. Conclusion

This study evaluated whether structured proxy features derived from baseline CT descriptors can improve multimodal NSCLC survival prediction. On Lung1, the primary four-modality benchmark achieved a C-index of 0.641, indicating improved benchmark-level performance relative to prior multimodal results (0.631 [15]). A separate exploratory coefficient-optimization analysis achieved a best observed C-index of 0.662, supporting the potential value of coefficient tuning while requiring further validation. The results suggest that compact rule-based proxy summaries may provide complementary prognostic information beyond independent deep, radiomic, and clinical descriptors within this benchmark framework. Within this pipeline, TMAE served as the imaging backbone after controlled comparison with alternative encoders. The framework uses only routinely acquired pretreatment CT and standard clinical variables, producing auditable proxy features without additional invasive assays or genomic profiling. The framework demonstrates retrospective prognostic ranking on a single public benchmark rather than clinically validated risk enrichment, surveillance planning, or treatment-related decision support. The most realistic next step is external multi-site validation with calibration, decision-threshold assessment, and prospective site-specific testing prior to any clinical deployment.

## Author Contributions

H.P.N.: Conceptualization, Methodology, Software, Validation, Formal analysis, Investigation, Writing - original draft, Writing - review and editing. D.H.: Writing - review and editing. E.S.: Writing - review and editing. Z.S.: Resources. J.Y.C.: Writing - review and editing, Supervision. All authors have reviewed the manuscript.

## Acknowledgements

This work was supported by the NIH Common Fund Data Ecosystem (CFDE) program under award U54OD036472, and by SPARC, UAB.

## Ethics Approval

This study used only de-identified retrospective imaging data from a publicly available repository (NSCLC-Radiomics / Lung1; DOI: 10.7937/K9/TCIA.2015.PF0M9REI), originally collected and approved by the MAASTRO Clinic Institutional Review Board, Netherlands, under the Cancer Genome Atlas (TCGA) protocol. Accordingly, this study was exempt from additional institutional review board approval.

## Consent to Participate

Informed consent was waived by the Institutional Review Board as this study used only de-identified retrospective imaging data from a publicly available repository (TCIA Lung1; DOI: 10.7937/K9/TCIA.2015.PF0M9REI), originally collected under the Cancer Genome Atlas (TCGA) protocol.

## Data and Code Availability

Imaging data and segmentations are publicly available from The Cancer Imaging Archive (NSCLC-Radiomics / Lung1; DOI: 10.7937/K9/TCIA.2015.PF0M9REI) [39]. Source code and documentation are available at: [https://github.com/aimed-lab/MLPA_paper](https://github.com/aimed-lab/MLPA_paper).

*Table 1. Simulation model notation.*

| Symbol | Definition | Units |
|---|---|---|
| Map(x) | CT tissue density proxy | [0, 1] |
| T, P, N, M | Viable core, proliferating rim, necrotic, malignant densities | cells $voxel^{-1}$ |
| $\rho$ | Total cell density, T + P + N + M | cells $voxel^{-1}$ |
| $\alpha$ | Proliferation rate | $\Delta t^{-1}$ |
| $\beta$ | Necrosis probability | $\Delta t^{-1}$ |
| $\mu$ | Maturation rate | $\Delta t^{-1}$ |
| $\gamma$ | Mutation rate | $\Delta t^{-1}$ |
| K | Voxel carrying capacity | cells $voxel^{-1}$ |
| $\tau_N$ | Necrosis onset threshold | $\Delta t$ |
| E, S | Tumor voxel intensity entropy; surface sphericity | — |

*Table 2. Cohort characteristics of the study population. Values are n (%) for categorical variables and median (IQR) for continuous variables.*

| Characteristic | Overall (N = 390) | Deceased (n = 345) | Alive (n = 45) | p |
|---|---|---|---|---|
| **Age, years** | | | | |
| Median (IQR) | 68.7 (61.5–75.8) | 69.4 (61.7–76.5) | 64.7 (59.8–70.7) | 0.002 |
| Mean ± SD | 68.1 ± 10.0 | 68.7 ± 10.1 | 64.0 ± 8.8 | |
| **Sex, n (%)** | | | | 0.064 |
| Male | 268 (68.7) | 243 (70.4) | 25 (55.6) | |
| Female | 122 (31.3) | 102 (29.6) | 20 (44.4) | |
| **Histology, n (%)** | | | | 0.716 |
| Squamous cell carcinoma | 144 (36.9) | 131 (38.0) | 13 (28.9) | |
| Large cell carcinoma | 108 (27.7) | 93 (27.0) | 15 (33.3) | |
| NOS | 56 (14.4) | 48 (13.9) | 8 (17.8) | |
| Adenocarcinoma | 49 (12.6) | 43 (12.5) | 6 (13.3) | |
| Unknown | 33 (8.5) | — | — | |
| **Overall stage, n (%)** | | | | 0.112 |
| Stage I | 83 (21.3) | 79 (22.9) | 4 (8.9) | |
| Stage II | 36 (9.2) | 32 (9.3) | 4 (8.9) | |
| Stage IIIA | 102 (26.2) | — | — | |
| Stage IIIB | 168 (43.1) | — | — | |
| Stage unknown | 1 (0.3) | — | — | |
| **Survival time, months** | | | | |
| Median (IQR) | 18.1 (9.0–48.8) | 16.1 (8.2–33.4) | 111.1 (59.5–119.3) | <0.001 |
| Deaths / events | 345 (88.5) | 345 (100.0) | 0 (0.0) | |

*IQR, interquartile range; NOS, not otherwise specified. p values: Mann-Whitney U test for continuous variables; Pearson chi-squared test for categorical variables. Dashes (—) indicate unavailable stratum-level counts.*

*Table 3. Model hyperparameters used across components.*

| Model | Key Hyperparameters |
|---|---|
| Neural Cox | Architecture: [32, 16] hidden units, batch normalization, ReLU, dropout = 0.30. Training: Adam, lr = 0.001, weight decay = $10^{-3}$, batch size = 16, max epochs = 200. Early stopping: patience = 20 epochs. |
| Transformer-based Masked Autoencoder | Input: $32^3$ patches. Tokenization: $4^3$ sub-patches, 512 tokens. Masking: 75%. Encoder: 768-dim, 12 layers, 12 heads. Decoder: 528-dim, 8 layers, 16 heads. Optimizer: AdamW (lr = $1.5\times10^{-4}$, weight decay = 0.05). Loss: MSE, Batch: 32, Epochs: 100. Output: 768-dim feature vector per patient (encoder CLS token). |
| Simulation framework | Maturation: 0.2. Mutation: 0.005. Iterations: 120. |

*Table 4. Comparison of NSCLC survival prediction methods on the TCIA Lung1 dataset. iAUC = integrated time-dependent AUC; — = not reported.*

| **Study** | **Method** | **C-index** | **iAUC** | **p-value** | **Ref** |
|---|---|---|---|---|---|
| Welch et al. | Cox + Radiomic Signature | 0.600 | — | — | [50] |
| Haarburger et al. | ResNet18 CNN + Radiomics | 0.623 | — | — | [51] |
| Haarburger et al. | Probabilistic U-Net + Radiomics | 0.577 | — | — | [52] |
| Scalco et al. | GP + Radiomics | 0.608 | — | — | [53] |
| Ferretti & Corino | Radiomics + AE + Clinical | 0.631 | 0.592 | 0.031 | [15] |
| Paolo et al. | EfficientNetB0 + Soft Attention | 0.584 | — | — | [54] |
| Our model (primary) | TMAE + SM + Radiomics + Clinical | 0.641 | 0.731 | <0.001 | — |
| **Our model (exploratory)** | **Coefficient optimization** | **0.662** | **0.748** | **<0.001** | **—** |

*Table 5. Sensitivity of simulation outputs to perturbations in proxy proliferation coefficient (α) and proxy necrosis coefficient (β). ΔRisk (abs) = mean absolute deviation of patient-level risk scores; Spearman ρ = rank correlation between perturbed and unperturbed patient risk rankings; Cat. Chg (%) = percentage of patients whose assigned risk category (high/low) changed; ΔC-index = change in concordance index relative to the unperturbed baseline.*

| Scenario | Δ (%) | ΔRisk (abs) | Spearman ρ | Cat. Chg (%) | ΔC-index |
|---|---|---|---|---|---|
| α +10% | +10 | 0.091 | 0.997 | 2.3 | −0.003 |
| β +10% | +10 | 0.024 | 0.998 | 2.3 | −0.002 |
| α + β +10% | +10 | 0.091 | 0.995 | 3.1 | −0.005 |
| α −10% | −10 | 0.073 | 0.997 | 3.8 | −0.004 |
| β −10% | −10 | 0.023 | 0.999 | 2.3 | −0.001 |
| α + β −10% | −10 | 0.078 | 0.996 | 4.6 | −0.002 |
| α +20% | +20 | 0.184 | 0.989 | 6.2 | −0.007 |
| β +20% | +20 | 0.030 | 0.998 | 3.1 | −0.003 |
| α + β +20% | +20 | 0.186 | 0.989 | 6.9 | −0.006 |
| α −20% | −20 | 0.135 | 0.989 | 7.7 | −0.001 |
| β −20% | −20 | 0.025 | 0.999 | 1.5 | −0.002 |
| α + β −20% | −20 | 0.134 | 0.990 | 7.7 | −0.001 |

*Table 6. Ablation studies. C = Clinical; R = Radiomics; TMAE = Transformer-based Masked Autoencoder; SM = Simulation. HR = hazard ratio; iAUC = integrated AUC. Each row reports the metrics from the best-performing model type for that feature configuration; therefore, HR values reflect model-specific risk-group separation and should be interpreted together with the corresponding C-index.*

| Configuration | Best Model | C-index | p-value | HR | iAUC |
|---|---|---|---|---|---|
| TMAE | CoxPH | 0.608 | 0.070 | 1.40 | 0.678 |
| TMAE + SM | Neural Cox | 0.636 | <0.001 | 2.31 | 0.713 |
| TMAE + R | Ensemble | 0.598 | 0.001 | 1.83 | 0.659 |
| TMAE + R + C | CoxPH | 0.600 | 0.018 | 1.55 | 0.654 |
| TMAE + SM + R + C | Ensemble | 0.641 | <0.001 | 1.95 | 0.731 |

*Table 7. Biological assumptions, rationale, and scope of each parameterization choice.*

| Assumption | Rationale | Scope |
|---|---|---|
| Higher entropy reflects greater intratumoral heterogeneity | Prior radiomics literature links entropy to intratumoral heterogeneity and worse NSCLC outcomes [5, 41, 43] | Not a direct measure of proliferation; used as a parameterization anchor |
| Lower sphericity reflects more irregular tumor morphology | Irregular shape is associated with invasive tumor structure and worse prognosis in NSCLC [42, 43] | Not a direct measure of necrosis; used as a parameterization anchor |
| Growth rate and necrosis ratio have prognostic value | Mitotic activity and tumor necrosis are established adverse prognostic indicators in NSCLC pathology [41, 42, 43] | The simulation provides imaging-derived surrogates, not histopathologically validated quantities |

*Table 8. Summary of the proposed framework: input, output, intended use, and constraints.*

| | |
|---|---|
| **Input** | Pretreatment CT volume, tumor segmentation mask (manual or automated), standard clinical variables (age, sex, stage, histology) |
| **Output** | Continuous relative risk score, exploratory dichotomized risk grouping for retrospective analysis |
| **Intended use** | Retrospective evaluation of baseline prognostic stratification from pretreatment CT |
| **Not intended for** | Treatment selection, replacement of clinician judgment, fixed treatment recommendations, or cross-site deployment without local recalibration and validation |

## Supplementary Figures

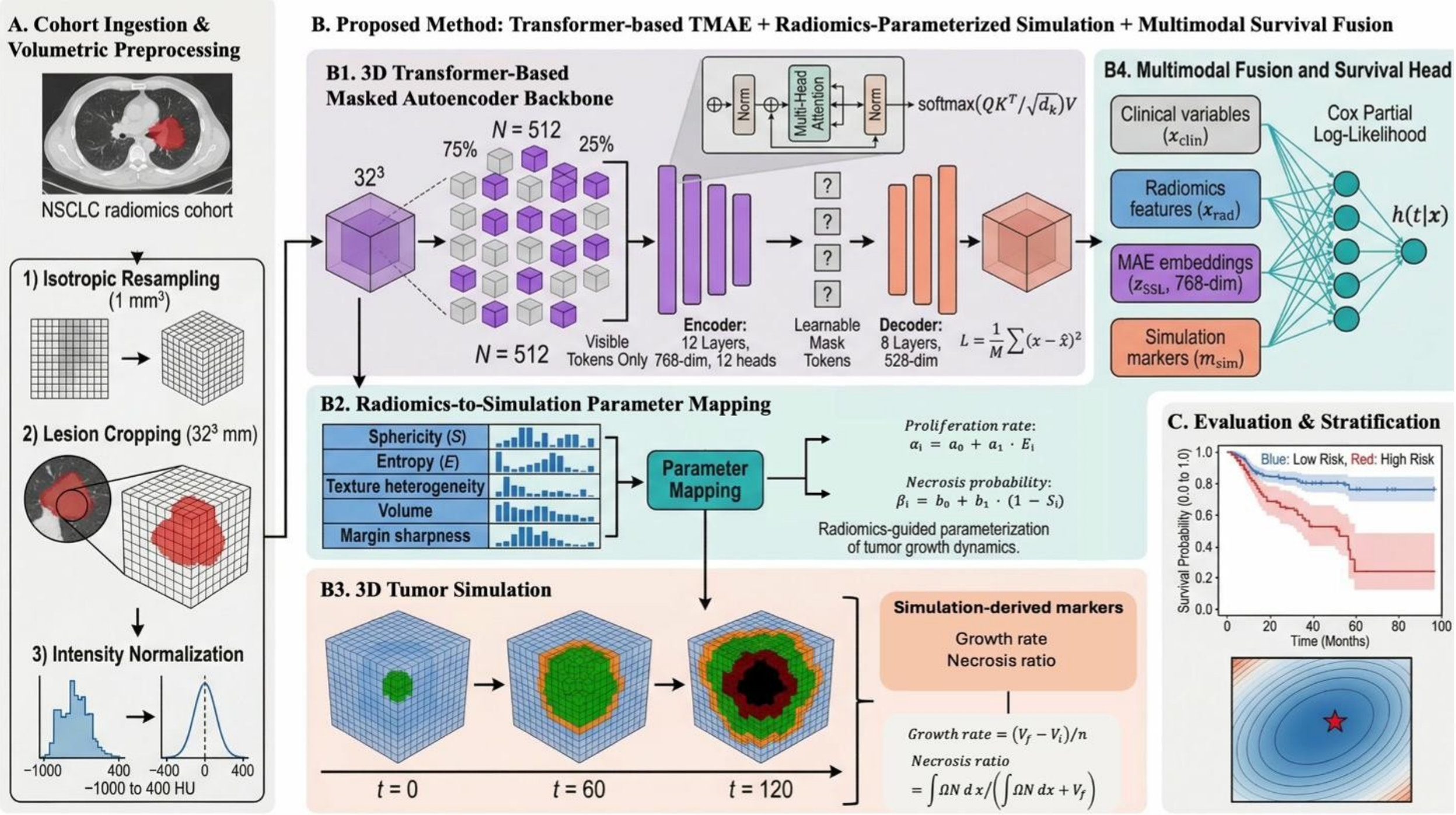

*Figure 1. Overview of the proposed pipeline for NSCLC survival prediction. (A) CT scans undergo isotropic resampling (1 mm³), lesion cropping (32³ mm), and intensity normalization. (B) The pipeline integrates three components: (B1) a TMAE (75% masking, 12-layer transformer encoder, 768-dim CLS token) for deep imaging embeddings; (B2) a Radiomic-to-Simulation Parameter Mapping panel depicting five radiomic descriptors (Sphericity, Entropy, Texture heterogeneity, Volume, and Margin sharpness), of which only Sphericity (S) and Entropy (E) serve as parameterization anchors for the cellular automaton, mapping to the proxy proliferation coefficient α and proxy necrosis coefficient β respectively; and (B3) a Simulation Framework yielding growth rate and necrosis ratio as simulation-derived markers. (B4) All four modalities fused under a Cox Partial Log-Likelihood objective. (C) Patients are stratified into low- and high-risk groups, evaluated by Kaplan–Meier analysis and C-index.*

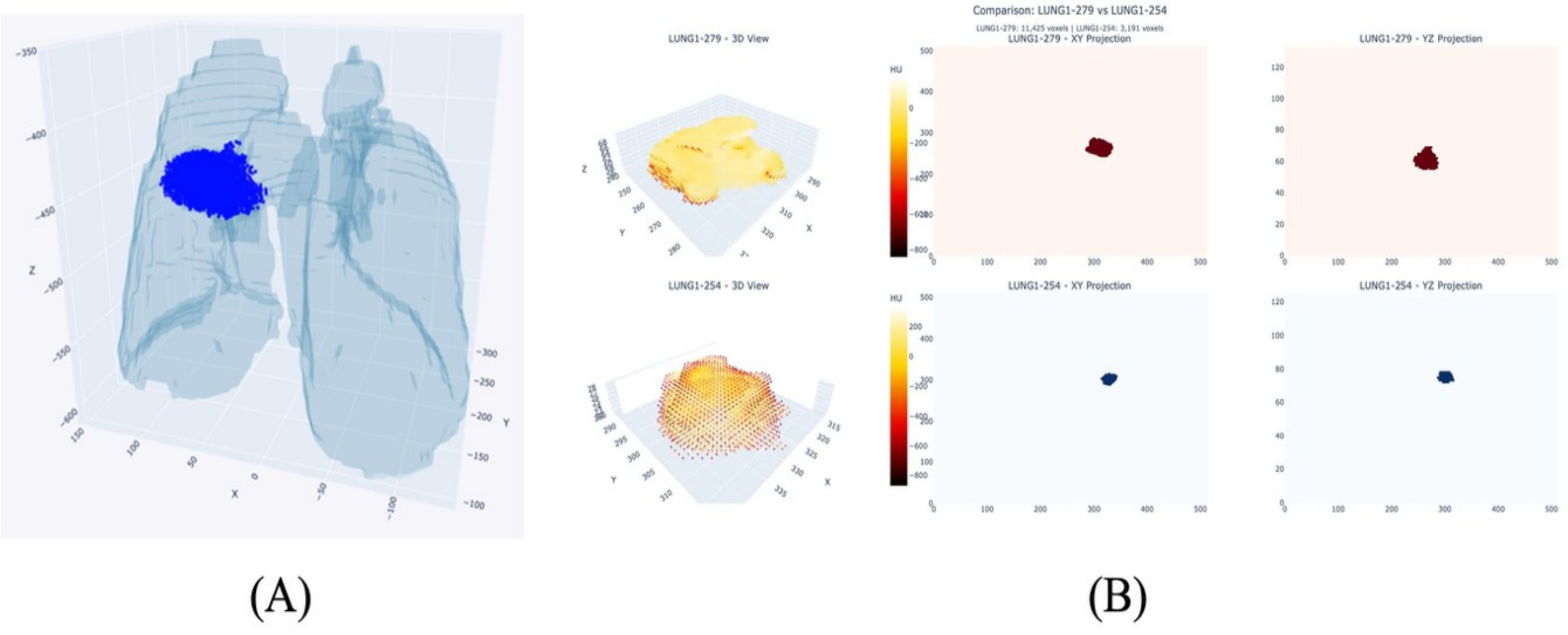

*Figure 2. Spatial tumor geometry and cross-patient size comparison in the Lung1 cohort. (A) Three-dimensional rendering showing lung volume (semi-transparent) and segmented primary tumor (solid blue voxel cloud). (B) Head-to-head comparison of LUNG1-279 vs LUNG1-254.*

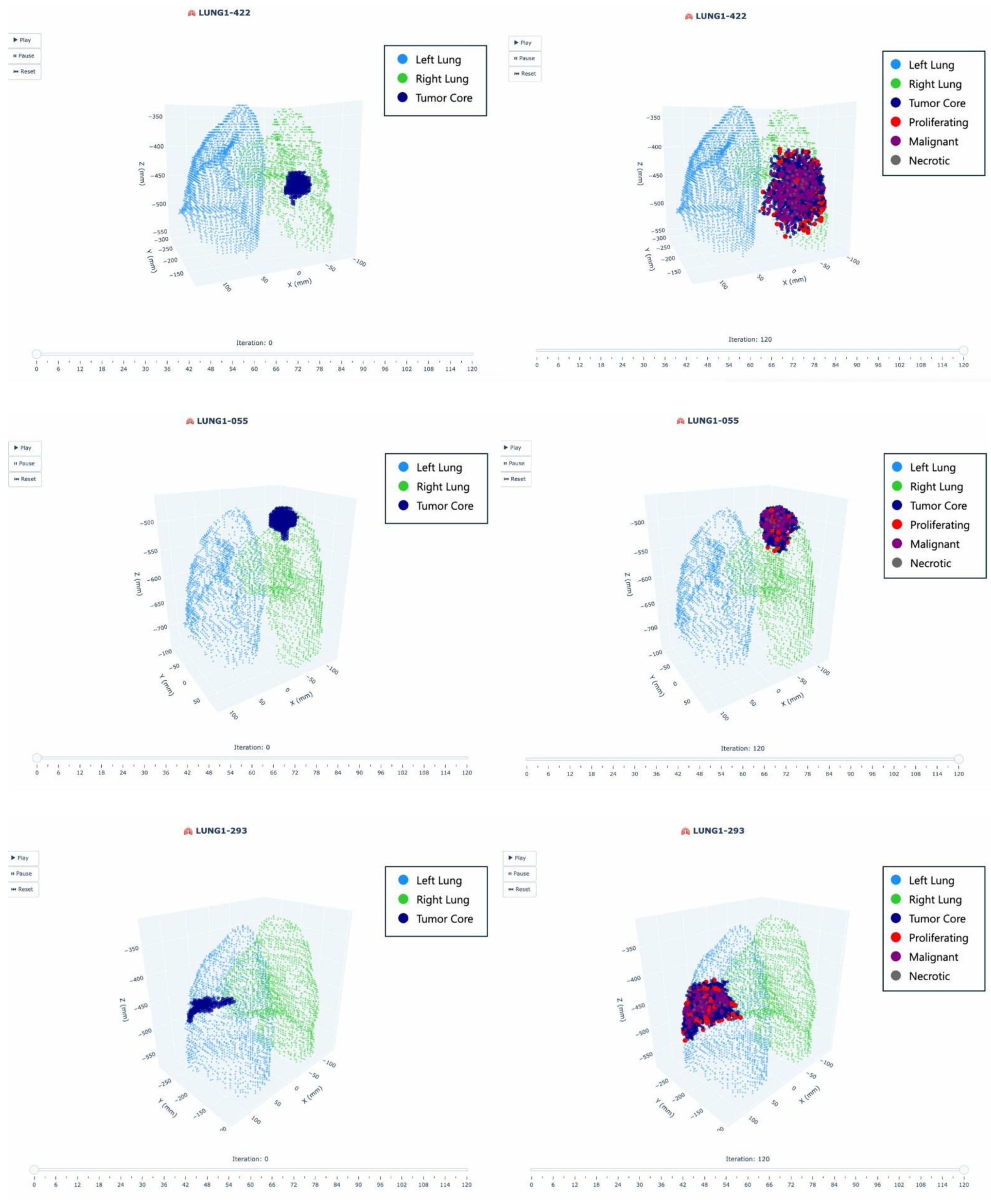

*Figure 3. Example 3D simulation growth dynamics. Three representative patients are shown at iteration 0 (left) and iteration 120 (right). Top: Lung1-422 ($\alpha = 0.0354$, $\beta = 0.0428$); Middle: Lung1-055 ($\alpha = 0.0168$, $\beta = 0.0410$); Bottom: Lung1-293 ($\alpha = 0.0230$, $\beta = 0.0630$). Cellular states: tumor core, proliferating, malignant, necrotic.*

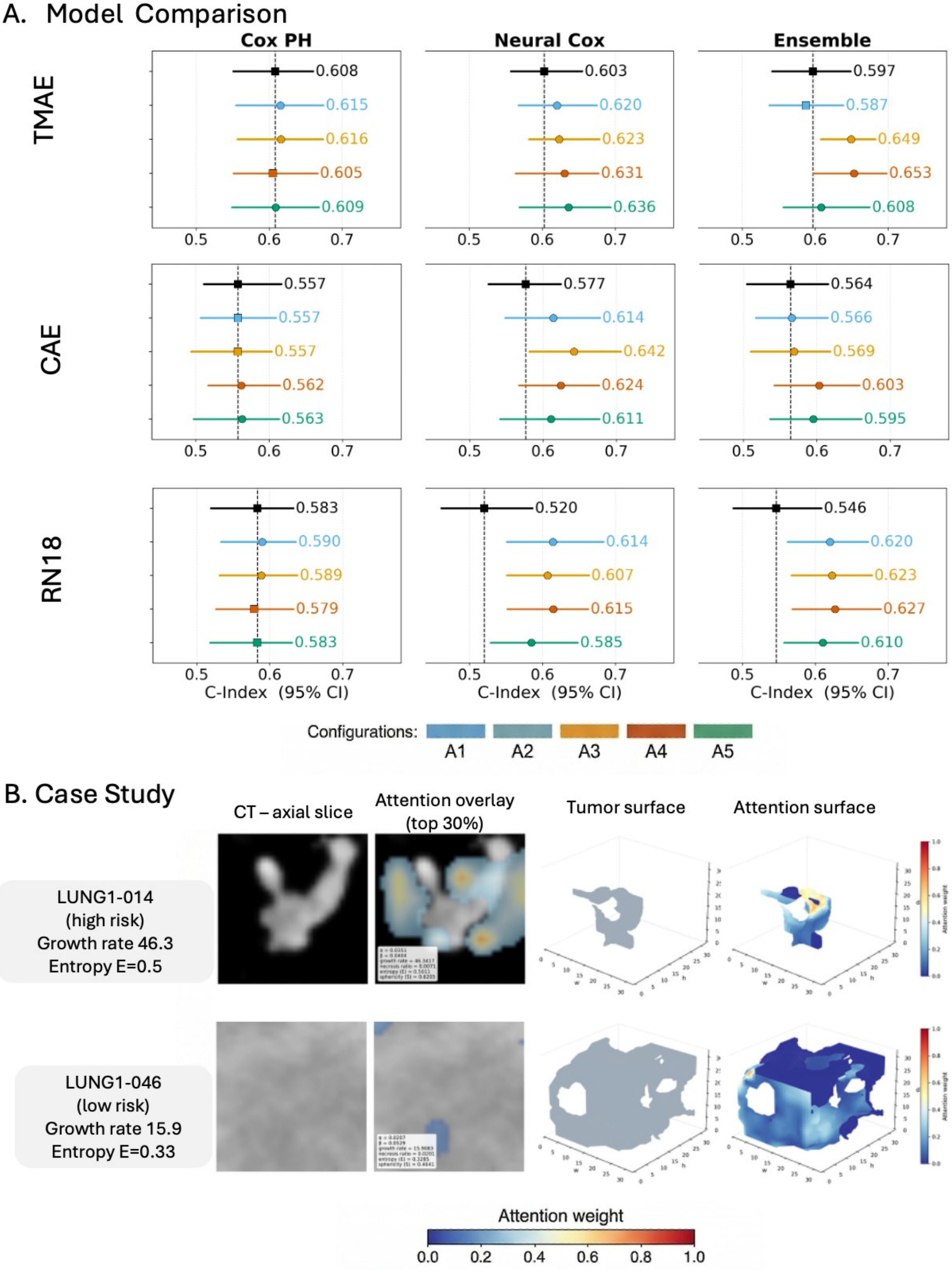

A. Model Comparison
Cox PH
Neural Cox
Ensemble
TMAE
0.608
0.615
0.616
0.605
0.609
0.603
0.620
0.623
0.631
0.636
0.597
0.587
0.649
0.653
0.608
CAE
0.557
0.557
0.557
0.562
0.563
0.577
0.614
0.642
0.624
0.611
0.564
0.566
0.569
0.603
0.595
RN18
0.583
0.590
0.589
0.579
0.583
0.520
0.614
0.607
0.615
0.585
0.546
0.620
0.623
0.627
0.610
0.5
0.6
0.7
C-Index (95% CI)
Configurations:
A1
A2
A3
A4
A5
B. Case Study
CT – axial slice
Attention overlay (top 30%)
Tumor surface
Attention surface
LUNG1-014
(high risk)
Growth rate 46.3
Entropy E=0.5
LUNG1-046
(low risk)
Growth rate 15.9
Entropy E=0.33
Attention weight
0.0
0.2
0.4
0.6
0.8
1.0

*Figure 4. (A) Comparative evaluation of imaging encoders across progressive simulation feature configurations. Forest plots show C-index with 95% CI for CoxPH, Neural Cox, and Ensemble models across five configurations: encoder-alone (A1), + radiomic scalars entropy and sphericity (A2), + proxy coefficients α and β (A3), + simulation outputs growth rate and necrosis ratio (A4), and + complete six-feature SM set (A5), for TMAE, CAE, and RN18. (B) CLS-to-patch attention maps for two representative test patients extracted from the final TMAE encoder block and averaged across 12 attention heads. Each row shows CT axial slice, attention overlay (top 30% patches by normalized attention weight), tumor surface, and attention surface. LUNG1-014 (high risk; growth rate 46.3; entropy E = 0.50) exhibits spatially structured attention concentrated at the morphological junctions of the branching tumor structure. LUNG1-046 (low risk; growth rate 15.9; entropy E = 0.33) exhibits near-uniform attention without any spatially identifiable high-attention region. TMAE = Transformer-based Masked Autoencoder; CAE = Convolutional Autoencoder; RN18 = MedicalNet-RN18; CLS = Classification Token; CT = Computed Tomography.*

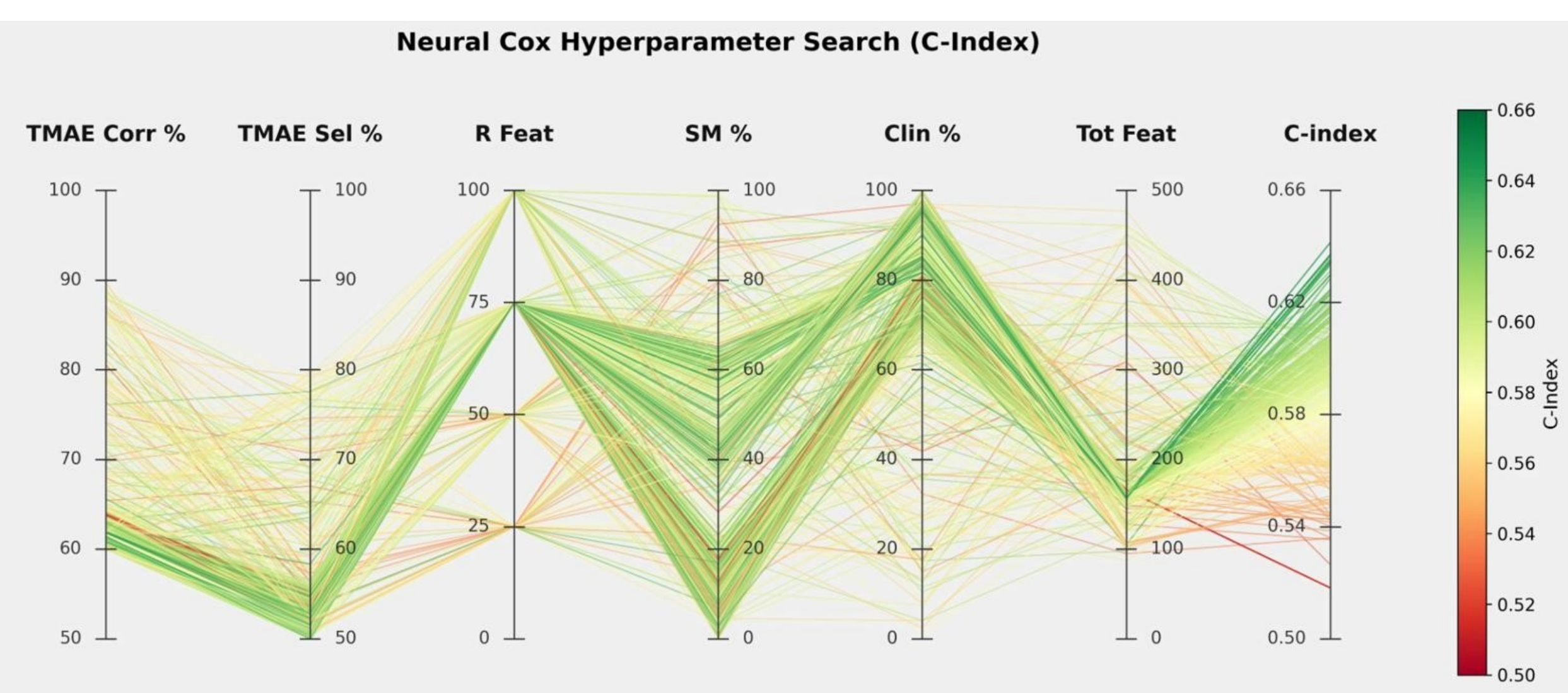

*Figure 5. Bayesian Hyperparameter Search. Parallel coordinates plot showing model performance (C-index) across five optimized hyperparameters (TMAE Corr %, TMAE Sel %, R Feat, SM %, Clin %) and one derived diagnostic quantity (Tot Feat) over 700 Bayesian TPE trials. Green lines indicate higher-performing configurations. TMAE = Transformer-based Masked Autoencoder; TPE = Tree-structured Parzen Estimator; R = Radiomics; SM = Simulation.*

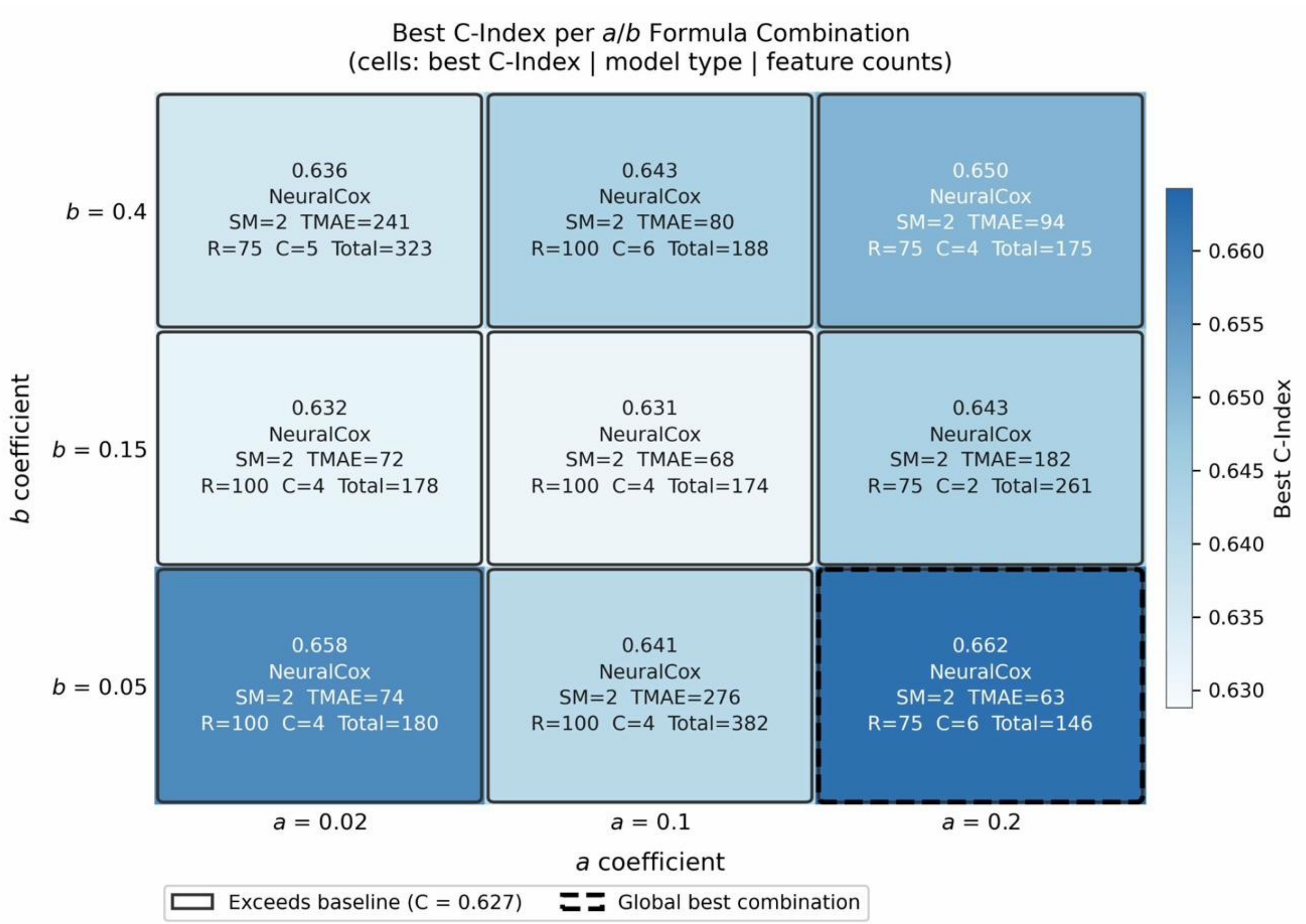

*Figure 6. Hyperparameter optimization results for the α/β parameter grid search. Heatmap of best C-index per α/β combination; each cell reports the C-index, best model type, and selected feature counts. Cells with solid borders exceed the internal RN18 baseline (C-index = 0.627); dashed border marks the global best combination (a = 0.2, b = 0.05, C-index = 0.662). TMAE = Transformer-based Masked Autoencoder; R = Radiomics; SM = Simulation; C = Clinical.*

## Supplementary Material: Implementation Details

This supplementary material provides the full implementation details for each experiment described in Section 4. Settings that are shared across all or most experiments are consolidated in Section S1 to avoid repetition; individual experiment sections (S2–S6) describe only the aspects that are specific to each experiment.

## S1. Common Experimental Settings

Model parameters and notation are summarized in Table 1; key hyperparameters across all components are listed in Table 2.

### *S1.1 Data Partitioning*

All experiments used the same fixed cohort split: patients were partitioned into a training set (2/3) and a held-out test set (1/3) using a stratified split by event status (random seed fixed at 42). All feature selection and model training was performed exclusively on the training partition.

### *S1.2 Modality Arms*

Four modality arms were used throughout all multimodal experiments:

- **Clinical (C):** Seven encoded covariates – age, sex, T-stage, N-stage, M-stage, overall stage, and histology.
- **Radiomics (R):** 1,688 normalized first- and higher-order imaging features derived from the tumor region.
- **TMAE:** 768-dimensional latent representations extracted from the [CLS] token of the self-supervised Vision Transformer encoder.
- **Simulation (SM):** Six parameters – tumor growth rate, necrosis ratio, proliferation coefficient α, necrosis coefficient β, radiomic entropy, and radiomic sphericity.

### *S1.3 Survival Models*

Three survival models were evaluated in all experiments:

- **Cox Proportional Hazards (CoxPH):** Fitted with LASSO penalization ($\lambda = 0.1$).
- **Neural Cox:** A feedforward neural survival model (see architecture below).
- **Ensemble:** Equal-weight combination of CoxPH and Neural Cox risk scores (weight = 0.5 each).

### *S1.4 Neural Cox Architecture*

The Neural Cox model comprised two fully-connected hidden layers of dimensions [32, 16] with batch normalization, ReLU activations, and dropout (rate = 0.30). Training used the Adam optimizer (learning rate = 0.001, weight decay = $10^{-3}$, batch size = 16) for up to 200 epochs, with early stopping (patience = 20) monitored on a 15% internal validation partition held out from the training set.

### *S1.5 Performance Metrics*

All experiments reported the following metrics on the held-out test set: concordance index (C-index) [46], integrated time-dependent AUC (iAUC) [47], log-rank p-value, and Cox-derived hazard ratio (HR) between median-split high- and low-risk groups.

### *S1.6 TMAE Feature Selection Pipeline*

Where TMAE feature selection was applied, a two-stage cross-validated pipeline was used exclusively on the training partition: (1) pairwise Pearson correlation filtering removing features with $|r|$ above a threshold $\rho$; (2) 10-fold cross-validation frequency filtering retaining only features selected in more than a threshold proportion $\tau$ of folds. The thresholds $\rho$ and $\tau$ were either optimized (Bayesian search experiments) or swept (ablation experiments), as described in the individual sections.

### *S1.7 Simulation Model Configuration*

The Simulation (SM) model is a 3D cellular automaton operating on a $90^3$ voxel grid over 120 iterations. The proliferation and necrosis parameters are derived from radiomics as $\alpha = a_0 + a_1 \cdot E$ and $\beta = b_0 + b_1 \cdot (1 - S)$, where E is radiomic entropy and S is sphericity, consistent with Equation (5). Simulation outputs, specifically tumor growth rate (net change in viable cell count per iteration) and necrosis ratio (fraction of necrotic cells at simulation end), are propagated as SM features into the downstream fusion model.

## S2. Comparative Evaluation of Imaging Models (Section 4.2)

Experiment-specific settings. Two self-supervised autoencoders and one pretrained CNN were compared within the same multimodal fusion framework, each producing a 768-dimensional output: (1) TMAE – a Vision Transformer trained via masked patch reconstruction (self-supervised), representation from the [CLS] token; (2) CAE – a Convolutional Autoencoder with a hierarchical strided-convolution encoder and fully-connected bottleneck; (3) RN18 – a 3D ResNet-18 pretrained on 23 medical imaging datasets, used as a fixed feature extractor with a linear projection head mapping 512-dimensional global average-pooled output to 768 dimensions (no task-specific fine-tuning). The five-arm progressive ablation described in Section 4.2 was applied to each encoder, with TMAE feature selection stabilized via a sweep over $\rho \in \{0.60, 0.70, 0.80, 0.90\} \times \tau \in \{0.50, 0.60, 0.70, 0.80\}$ and the threshold maximizing Neural Cox C-index selected for final reporting.

## S3. Sensitivity Analysis of Simulation Parameters (Section 4.4.2)

Experiment-specific settings. Twelve perturbation scenarios were evaluated by applying independent ±10% and ±20% relative changes to the proliferation coefficient $\alpha$, the necrosis coefficient $\beta$, and both simultaneously. For each scenario, the Simulation model (Section S1.7) was re-executed from scratch for all test patients using the perturbed parameter values; patients for whom CT data could not be loaded retained their original simulation outputs. CT volumes were pre-loaded in parallel using multithreaded joblib workers. Feature selection was fixed at the optimal configuration from Section 4.4: TMAE thresholds $\rho = 0.6162$, $\tau = 0.5361$ over 10 folds; top 75 Radiomics features by univariate Cox ranking; top 67.51% of Clinical features by univariate Cox p-value; all six SM features retained without filtering. The survival model was CoxPH only (LASSO $\lambda = 0.1$), trained on the fixed 2/3 training partition.

## S4. Bayesian Feature Selection and Model Optimization (Section 4.4.1)

Experiment-specific settings. Bayesian optimization was performed using the Tree-structured Parzen Estimator sampler with 700 total trials and 140 random startup trials. The objective function returned the maximum C-index achieved across the three survival models per trial. Five hyperparameters were jointly optimized: the TMAE correlation threshold $\rho \in [0.60, 0.90]$, the TMAE selection frequency threshold $\tau \in [0.50, 0.80]$, the number of top Radiomics features $k \in \{25, 50, 75, 100\}$ (ranked by univariate Cox p-value), the SM feature retention fraction $\pi SM \in [0.0, 1.0]$, and the Clinical feature retention fraction $\pi C \in [0.0, 1.0]$. Feature selection was executed once per trial on the training partition and the resulting feature matrix passed to all three survival models on the same held-out test set.

## S5. Multimodal Feature Fusion Ablation Study (Section 4.3)

Experiment-specific settings. A total of 18 configurations were evaluated: four single-domain baselines, seven feature-level fusion combinations (concatenated features prior to model training), and seven signature-level fusion combinations (per-domain CoxPH risk scores concatenated as second-stage inputs). For the TMAE and Radiomics arms, feature selection followed the common two-stage pipeline (Section S1.6) with fixed thresholds: Pearson $|r| > 0.80$ for correlation filtering, and 70% cross-validation selection frequency (10-fold LASSO Cox filter). Radiomics features were further reduced to the top 10 by univariate Cox p-value; TMAE to the top 15. Clinical and SM features were used without cross-validation-based filtering.

## S6. Optimization of Simulation Coefficients for the Proposed Framework (Section 4.4.3)

Experiment-specific settings. A grid search was conducted over nine ($\alpha$, $\beta$) formula combinations: $a \in \{0.02, 0.10, 0.20\}$ and $b \in \{0.05, 0.15, 0.40\}$. For each combination, the Simulation model (Section S1.7) was re-executed on the full cohort. A 100-trial Bayesian TPE optimization (20 random startup trials) was run per combination, jointly optimizing four hyperparameters: TMAE $\rho \in [0.60, 0.90]$, TMAE $\tau \in [0.50, 0.80]$, Radiomics $k \in \{25, 50, 75, 100\}$, and Clinical retention $\pi C \in [0.30, 1.00]$. The SM arm was fixed at two outputs (growth rate and necrosis ratio) without selection. The Ensemble model weighted CoxPH and Neural Cox equally (gradient boosting weight = 0). All nine combinations were processed in parallel with checkpoint-and-resume support.